\documentclass{IOS-Book-Article}

\usepackage{mathptmx}
\usepackage{soul}\setuldepth{article}
\usepackage{bera}
\usepackage{listingsutf8}
\usepackage{xcolor}
\usepackage[utf8]{inputenc}
\usepackage{graphicx}
\usepackage{url}

\usepackage{mathptmx}
\usepackage{tabularx}
\usepackage{times}
\usepackage{latexsym}
\usepackage{layout}
\usepackage{float}
\usepackage{placeins}
\usepackage{wrapfig}
\usepackage{afterpage}
\usepackage[normalem]{ulem}
\usepackage{color}
\usepackage{nicefrac}
\usepackage{multirow}
    
\colorlet{punct}{red!60!black}
\definecolor{background}{HTML}{EEEEEE}
\definecolor{delim}{RGB}{20,105,176}
\colorlet{numb}{magenta!60!black}

\lstdefinelanguage{json}{
    basicstyle=\normalfont\ttfamily,
    numbers=left,
    numberstyle=\scriptsize,
    stepnumber=1,
    numbersep=8pt,
    showstringspaces=false,
    breaklines=true,
    frame=lines,
    backgroundcolor=\color{background},
    literate=
     *{0}{{{\color{numb}0}}}{1}
      {1}{{{\color{numb}1}}}{1}
      {2}{{{\color{numb}2}}}{1}
      {3}{{{\color{numb}3}}}{1}
      {4}{{{\color{numb}4}}}{1}
      {5}{{{\color{numb}5}}}{1}
      {6}{{{\color{numb}6}}}{1}
      {7}{{{\color{numb}7}}}{1}
      {8}{{{\color{numb}8}}}{1}
      {9}{{{\color{numb}9}}}{1}
      {:}{{{\color{punct}{:}}}}{1}
      {,}{{{\color{punct}{,}}}}{1}
      {\{}{{{\color{delim}{\{}}}}{1}
      {\}}{{{\color{delim}{\}}}}}{1}
      {[}{{{\color{delim}{[}}}}{1}
      {]}{{{\color{delim}{]}}}}{1}
    {ā}{{\={a}}}1
    {Ā}{{\={A}}}1
    {Č}{{\'c}}1
    {Ć}{{\'{C}}}1
    {ē}{{\={e}}}1
    {Ē}{{\={E}}}1
    {ģ}{{\v{g}}}1
    {Ģ}{{\v{G}}}1
    {ī}{{\={i}}}1
    {Ī}{{\={I}}}1
    {ķ}{{\v{k}}}1
    {Ķ}{{\v{K}}}1
    {ļ}{{\c{l}}}1
    {Ļ}{{\c{L}}}1
    {ń}{{\c{n}}}1
    {Ń}{{\c{N}}}1
    {š}{{\v{s}}}1
    {Š}{{\v{S}}}1
    {ū}{{\={u}}}1
    {Ū}{{\={U}}}1
    {ž}{{\v{z}}}1
    {Ž}{{\v{Z}}}1,
}

\def\hb{\hbox to 10.7 cm{}}

\begin{document}

\pagestyle{headings}
\def\thepage{}

\begin{frontmatter}              

\title{What Can We Learn From Almost a Decade of Food Tweets}

\markboth{}{April 2020\hb}

\author[A]{
    \fnms{Uga} 
    \snm{Sproģis}%
    \thanks{Corresponding Author: Uga Sproģis; E-mail: ugasprogis12@inbox.lv.}
},
\author[B]{
    \fnms{Matīss} 
    \snm{Rikters}
}

\runningauthor{Sproģis & Rikters}
\address[A]{Faculty of Computing, University of Latvia}
\address[B]{The University of Tokyo}

\begin{abstract}
    We present the Latvian Twitter Eater Corpus - a set of tweets in the narrow domain related to food, drinks, eating and drinking. The corpus has been collected over time-span of over 8 years and includes over 2 million tweets entailed with additional useful data. We also separate two sub-corpora of question and answer tweets and sentiment annotated tweets. We analyse contents of the corpus and demonstrate use-cases for the sub-corpora by training domain-specific question-answering and sentiment-analysis models using data from the corpus.
\end{abstract}

\end{frontmatter}
\markboth{April 2020\hb}{April 2020\hb}

\section{Introduction}

Even though usage and popularity of Twitter have stopped rapidly growing and even dropped in recent years\footnote{https://www.statista.com/statistics/282087/number-of-monthly-active-twitter-users}, it still has a considerable amount of loyal users who keep on sharing everything from worldwide events to random personal details with their followers. We decided to focus on one of the random personal details that people share, specifically - anything to do with food consumption and related topics. 

Several corpora of Latvian tweets exist in prior work, but none of them are domain-specific and have been collected over an extensive period of time.
Milajevs \cite{milajevs2018language} collected and analysed 1.4 million tweets geo-located in Riga, Latvia from April 2017 to July 2018 and 60 thousand tweets \cite{milajevs-2017-toward} from November 2016 to March 2017. Pinnis \cite{pinnis2018latvian} collected and analysed 3.8 million tweets of Latvian politicians, companies, media, and users who interacted with these entities from August 2016 to July 2018 There are also several data sets of general sentiment-annotated tweets \cite{peisenieksuses,viksna2018sentiment,pinnis2018latvian}\footnote{https://github.com/nicemanis/LV-twitter-sentiment-corpus} amounting to 14,781 tweets in total. 

In this paper, we describe the Twitter eater corpus (TEC) and analyse its contents. We also provide two sub-corpora - one consisting of question and answer tweets and one with sentiment-annotated tweets. More details can be found in Section \ref{sec:corpus}. In Sections \ref{sec:qa-exp} and \ref{sec:sa-exp} we describe question answering and sentiment analysis experiments using our corpus. Finally, we conclude the paper in Section \ref{sec:conclusion}. 

\section{The Twitter Eater Corpus}
\label{sec:corpus}

The corpus consists of tweets that have been collected from October 2011 \cite{rikters2012universalas} until April 2020. They are tracked using 363 keywords, which are various inflections of Latvian words associated with eating, tasting, breakfast, lunch, dinner, etc. The main keywords are shown in Table \ref{tab:main-keywords} - the words in bold are mostly verbs that describe eating - these were inflected to all usable forms and included in the full keyword list. The rest of the keywords are a set of the top 60 food-related words that were most popular in the first month of collecting the tweets.

Figure \ref{fig:tweet-example} illustrates the contents of a single tweet from the TEC in JSON notation. Each tweet consists of primary fields - \textit{"tweet\_id"}, \textit{"tweet\_text"}, \textit{"tweet\_author"} and \textit{"created\_at"}, which will always be present, and optional fields, which depend on the tweet text and metadata. We separate three groups of optional fields: 1) "\textit{media\_url}" and \textit{"expanded\_url"}, which contain information about media files from the tweet; 2) \textit{"location\_name"}, \textit{"location\_lng"}, \textit{"location\_lat"} and \textit{"location\_country"}, which specify where the tweet was created; and 3) \textit{"food\_surface\_form"}, \textit{"food\_nominative\_form"}, \textit{"food\_group"} and \textit{"food\_english\_translation"}, which contain semicolon-separated lists of foods or drinks that appear in the tweet.

At the beginning of the project approximately 15,000 food and drink words from collected tweets were manually annotated with their respective nominative forms, English translations and food groups according to the food guide pyramid \cite{duston_1992}. The food groups are: bread, cereal, rice, pasta (6); vegetables (5); fruit, berries (4); milk products (3); meat, eggs, fish (2); fats, oils, sweets (1). There are two additional groups for drinks - alcoholic drinks (7) and non-alcoholic drinks (8).

The corpus is available on Github\footnote{https://github.com/Usprogis/Latvian-Twitter-Eater-Corpus} in accordance with the content redistribution section of the Twitter Developer Agreement and Policy\footnote{https://developer.twitter.com/en/developer-terms/agreement-and-policy}. The public release includes tweet IDs along with data fields created within the scope of this project (starting with \textit{"location\_lng"} in Figure \ref{fig:tweet-example}). The complete version is available upon individual request for research purposes. The repository also includes data processing scripts and details on how to reproduce our experiments.

\begin{table}[b]
    \caption{List of main keywords used to collect the corpus.}
    \begin{tabular}{llllll}
    \hline
    \textbf{taste} & \textbf{lunch} & beet & potato & mandarin & sweet \\ 
    \textbf{eat} & \textbf{feast} & bun & cabbage & sauce & mushroom \\ 
    \textbf{breakfast} & \textbf{drink} & carrot & candy & pancake & onion \\ 
    \textbf{dine} & \textbf{treat} & chips & sour cream & dumpling & chocolate \\ 
    \textbf{dinner} & \textbf{nom} & vegetable & cream soup & gingerbread & tea \\ 
    \textbf{bite} & \textbf{appetite} & meat & cake & rice & tomato \\ 
    \textbf{meal} & orange & Hesburger & drink & salad & grape \\ 
    \textbf{food} & apple & coffee & McDonald's & ice cream & strawberry \\ \hline
    \end{tabular}
  \label{tab:main-keywords}
\end{table}

\begin{figure}[t]
    \begin{small}
    \begin{lstlisting}[language=json,numbers=none]
    {
      "tweet_id":  1213025400273735680,
      "tweet_text":  "Gulašzupa #receptesĪsumā gulašzupa ir gana vienkārša liellopu gaļas bāzēta zupa https://t.co/OnqDwotQr0 https://t.co/Z2tAodyj9M",
      "tweet_author":  "receptes_eu",
      "created_at":  "2020-01-03 11:12:54",
      "media_url":  "http://pbs.twimg.com/media/ENWIKb8WsAAiLKE.jpg",
      "expanded_url":  "https://twitter.com/receptes_eu/status/1213025400273735680/photo/1",
      "location_name":  "Ogresgals",
      "location_lng":  "24.7377",
      "location_lat":  "56.8079",
      "location_country":  "Latvia",
      "food_surface_form":  "Gulašzupa;liellopu;gaļas;zupa;",
      "food_nominative_form":  "gulašs;liellops;gaļa;zupa;",
      "food_group":  "2;2;2;6;",
      "food_english_translation":  "Goulash;Cattle;Meat;Soup;",
    }
    \end{lstlisting}
    \end{small}
  \caption{An example of a tweet from the TEC with all available metadata.}
  \label{fig:tweet-example}
\end{figure}


\begin{table}[b]
    \caption{List of foods and drinks which are the most popular overall.}
    \begin{tabular}{lrlr}
    \hline
    \textbf{Food} & \textbf{Count} & \textbf{Drink} & \textbf{Count}  \\ \hline 
    Chocolate & 117,235 & Tea & 163,338 \\ 
    Ice cream & 86,109 & Coffee & 120,040 \\ 
    Meat & 85,574 & Juice & 18,179 \\ 
    Potatoes & 70,135 & Water & 15,692 \\ 
    Salads & 61,616 & Beer & 14,845 \\ 
    Cake & 52,267 & Cocktails & 8,207 \\
    Soup & 46,545 & Coca-cola & 5,016 \\
    Pancakes & 40,203 & Alcohol & 4,766 \\
    Sauce & 40,201 & Champagne & 3,673 \\
    Apple & 36,571 & Vodka & 2,802 \\ \hline
    \end{tabular}
  \label{tab:top-foods-drinks}
\end{table}


\subsection{Content Overview}

The corpus contains 2,275,787 tweets, of which 155,057 contain media information, 165,335 contain location information and 1,297,159 tweets mention foods or drinks.
Table \ref{tab:top-foods-drinks} shows the 10 most popular foods and drinks from the TEC. Looking from a Latvian consumer perspective\footnote{https://enciklopedija.lv/skirklis/4980-nacion\%C4\%81l\%C4\%81-virtuve-Latvij\%C4\%81} it is very typical that Latvians mostly drink water, tea, juice, beer and eat meat, vegetables and fruits. Interesting, however, is the high popularity of sweets such as chocolate, cakes, ice cream and Coca-Cola.

\begin{figure}[b]
    \includegraphics[width=\linewidth]{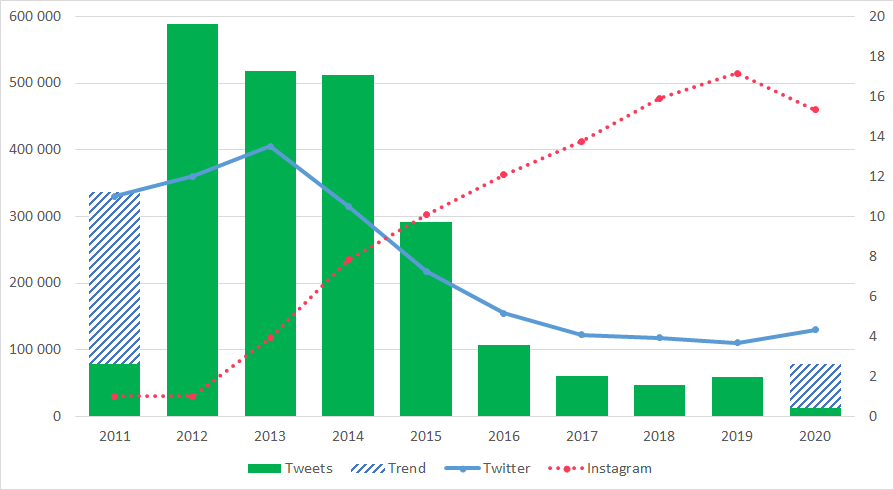}
    \caption{Collected tweet count by year.}
    \label{fig:yearly-counts}
\end{figure}

Figure \ref{fig:yearly-counts} shows the yearly count of collected tweets along with the potential trend (since for years 2011 and 2020 only a part has been collected) and the general popularity of Twitter and Instagram (a competing social network) for Latvia from Google Trends \footnote{https://trends.google.com/trends/explore?hl=en-US\&tz=-540\&date=2011-10-06+2020-03-14\&geo=LV\&q=\%2Fm\%2F0fjd36,\%2Fm\%2F0289n8t,\%2Fm\%2F02y1vz,\%2Fm\%2F0glpjll\&sni=3}. There was a stable income of food tweets up until 2015, but after that, it seems that the decrease correlates with the overall drop in popularity of Twitter in Latvia, which seems to be directly opposite to the popularity of Instagram in Latvia according to Google Trends.

\begin{figure}[t]
    \includegraphics[width=\linewidth]{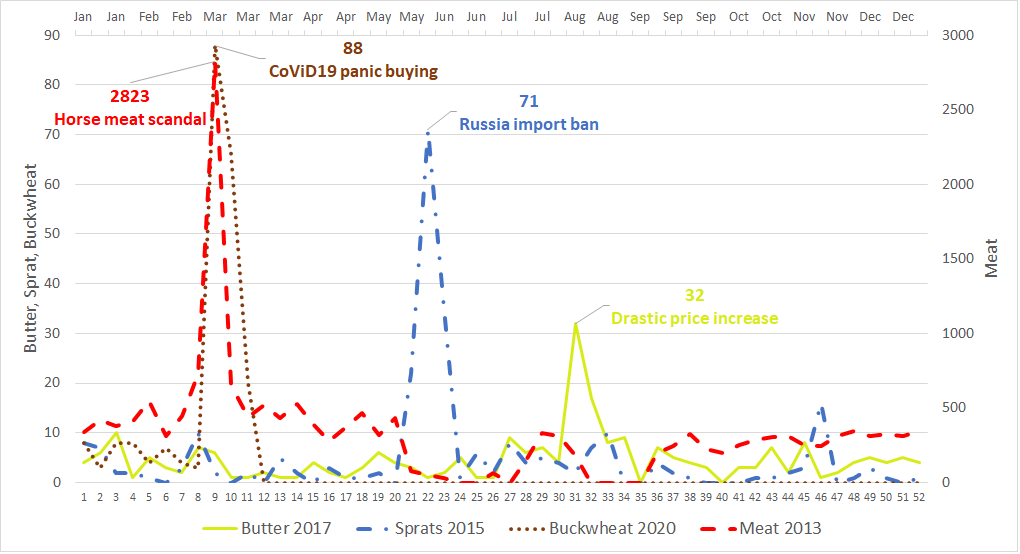}
    \caption{Four of the large trends noticeable in the TEC.}
    \label{fig:big-trends}
\end{figure}

In Figure \ref{fig:big-trends} we have visualised four of the largest tweet trends over the past years from the Latvian speaking twitter users. The most recent one just a month ago - panic buying of buckwheat due to the CoViD19 pandemic of 2020, followed by the doubling of butter prices in 2017, Latvian sprat import ban to Russia in 2015, and finally the horsemeat scandal in 2013. If we look closer at the 2823 tweets about meat in week 9 of 2013, we can see multiple inflexions of the word "horse" along with words like "scandal" and "investigation" among the most common words.

\begin{figure}[t]
    \includegraphics[width=\linewidth]{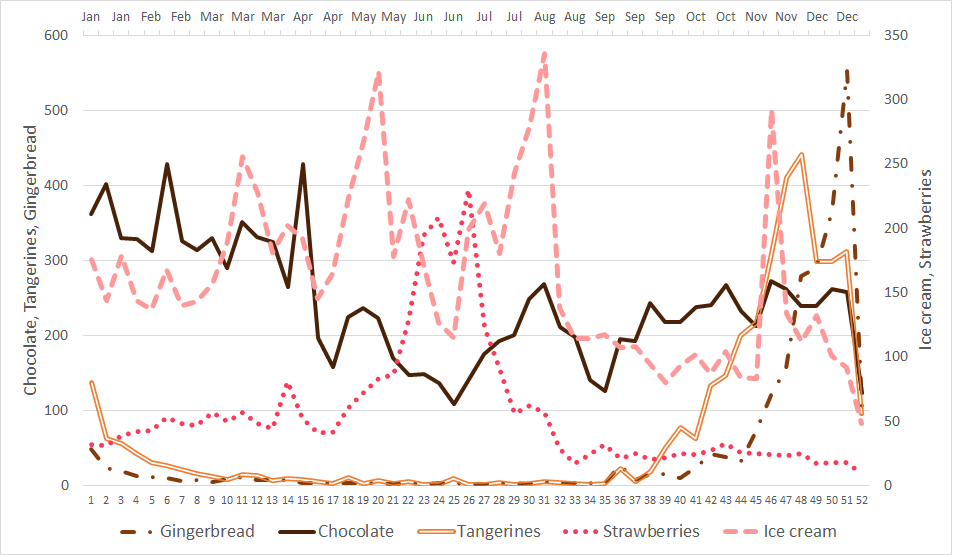}
    \caption{Five of the yearly seasonal trends noticeable in the TEC.}
    \label{fig:seasonal-trends}
\end{figure}

Figure \ref{fig:seasonal-trends} shows a selection of seasonal trends averaged from data between 2012 and 2019. Most trends have one peak zone indicating parts of the year when they are more popular. Examples of this are gingerbread and tangerines in December, and strawberries and ice cream in the summer. We were expecting to see chocolate peak high on Valentine's day, but while it does peak, the difference is not as high.

\subsection{Question - Answer Sub-corpus}
We noticed that there are plenty of tweets in our corpus that express questions. To highlight one of the uses of the corpus, we selected a subset of tweets which include at least one of typical Latvian question words\footnote{http://valoda.ailab.lv/latval/vidusskolai/SINTAKSE/sint3jaut.htm} or phrases along with a question mark. This resulted in 215,233 question tweets. To gather answers for them, we scraped Twitter's web version\footnote{https://github.com/luodaoyi/TwEater}, which resulted in 19,871 tweets with at least one reply. Since there were many tweets with multiple answers, we eventually wound up with 42,744 question-answer pairs. We randomly selected subsets of 1000 and 500 question-answer pairs to use as the development set and evaluation set respectively.

\subsection{Sentiment Annotated Sub-corpus}
We manually annotated 5420 tweets. marking them as positive, neutral or negative. This gave us 1631 positive, 2507 neutral and 1282 negative tweets. We further split these into a test set of 250 tweets from each class and a training set 

\section{Experiments}
\subsection{Question Answering}
\label{sec:qa-exp}
Typical question answering systems are trained using paragraphs of text, questions about the paragraphs and answers to those questions \cite{rajpurkar-etal-2016-squad}. Since we only had question-answer, we chose to train an encoder-decoder model similar to machine translation using questions and answer as source and target languages respectively. We used Sockeye \cite{Sockeye:17} to train transformer architecture models with the base parameters until they reached convergence on development data.

Our initial experiments using only TEC data showed rather poorly generated answers due to lack of general-domain training data. To mitigate this, we used the same approach to select question-answer tweets from the Latvian Tweet Corpus \cite{pinnis2018latvian}. This gave us 546,982 additional question-answer pairs to add to our training data.

\subsubsection{Data Pre-processing}
We performed tokenisation or truecasing using scripts from the Moses Toolkit \cite{Koehn2007Moses:Translation}. We used Sentencepiece \cite{kudo2018sentencepiece} to create a shared subword vocabulary of 8000 tokens. We replaced all Twitter-specific \textit{@user} mentions with \textit{@USR} and URLs with \textit{@URL}, as these usually don't contain relevant linguistic data for the model to learn. We also replaced multiple consecutive \textit{@USR} or \textit{@URL} tags with a single one and removed them completely if they were either at the start of the tweet or in the end.

\subsubsection{Results}

Figure \ref{fig:qa-examples} shows several examples of generated answers to the evaluation data questions. There were many hypothesis answers that were not even close to the reference ones but still made some sense in relation to the questions, such as the first two. There were also just as many or even more answers that made no sense at all like the last one.

We performed a small-scale human evaluation on the results by asking 5 annotators to evaluate a random 10\% of the evaluation set by marking generated answers as either OK or not good (NG). The evaluators marked 46.40\% of answers as OK. The evaluators had an overall agreement of 66.27\% (Free-marginal kappa \cite{randolph2005free} - 0.33), which indicates moderate agreement.

\begin{figure}[t]
  \begin{tabular}{lp{10.4cm}}
   \bf Question: & Bet es jau tevis dēļ visu daru. Cikos jāiet ēst torti? \\
   \bf Hypothesis: & Es jau paēdu.: D \\
   \bf Reference: & uzzvanīšu kad varēs nākten ciemen. \\ \\
   
   \bf Question: & Man šķūnī nesen piedzima ežuļi. Tāpēc SOS jautājums - ko ēd mazi eži? \\
   \bf Hypothesis: & Jā, tas ir ļoti labs jautājums.:) \\
   \bf Reference: & tikai nebaro ar krūti, viņi nepanes pienu! \\ \\
   
   \bf Question: & kāpostu tīteņi vai cepelīni? \\
   \bf Hypothesis: & Jā, svaigēdājus. \\
   \bf Reference: & kāpostu tīteņi \\
  \end{tabular}
  \caption{Example questions with generated hypothesis answers and true reference answers.}
  \label{fig:qa-examples}
\end{figure}

\subsection{Sentiment Analysis}
\label{sec:sa-exp}

For the sentiment analysis experiment, we used 3116 of our sentiment-annotated tweets. 250 random tweets from each sentiment class were separated as an evaluation corpus and the remaining 2368 were used for training. 

For sentiment analysis, we performed similar data pre-processing steps as for question answering, except for splitting words in sentencepieces. We also experimented with stemming\footnote{https://github.com/rihardsk/LatvianStemmer} and lemmatizing \cite{Paikens:2007:BalticHLT} words.

Table \ref{tab:sentiment-experiment} shows results of our sentiment analysis experiments. We compared a Python implementation of the Naive Bayes classifier from NLTK\cite{bird2009natural} against Pinnis \cite{pinnis2018latvian} implementation of the Perceptron classifier. We also experimented with several combinations of training data sets - TE (our Twitter Eater dataset), MP \cite{pinnis2018latvian}, RV \cite{viksna2018sentiment}, PE \cite{peisenieksuses}, NI\footnote{https://github.com/nicemanis/LV-twitter-sentiment-corpus}. We found that the highest classification accuracy - 61.23\% - is achieved by using all but NI data sets for training and only stemming all words.

\begin{table}[b]
    \begin{tabular}{lcccccc}
    \\ \hline
    \textbf{Training   Data} & \textbf{TE} & \textbf{MP} & \textbf{MP.PE} & \textbf{TE.MP} & \textbf{All} & \textbf{TE.MP.RV.PE} \\ \hline
    \textbf{Naive Bayes}     & 53.21       & 43.32       & 45.72          & 56.55          & 59.63                   & 58.02                \\
    \textbf{Perceptron}      & 53.07       & 52.67       & 53.47          & 57.87          & 57.33                   & 58.27                \\ \hline
                             & \multicolumn{6}{c}{\textbf{Stemmed}}                                                                         \\ \hline
    \textbf{Naive Bayes}     & 53.74       & 46.39       & 50.67          & 58.16          & 60.56                   & \textbf{61.23}       \\
    \textbf{Perceptron}      & 56.67       & 53.73       & 54.13          & 60.00          & 56.93                   & 57.73                \\ \hline
                             & \multicolumn{6}{c}{\textbf{Lemmas}}                                                                          \\ \hline
    \textbf{Naive Bayes}     & 53.88       & 45.45       & 49.60          & 56.42          & 58.42                   & 59.63                \\
    \textbf{Perceptron}      & 54.41       & 51.07       & 53.07          & 57.35          & 56.95                   & 56.95                \\ \hline
                             & \multicolumn{6}{c}{\textbf{Stemmed Lemmas}}                                                                  \\ \hline
    \textbf{Naive Bayes}     & 54.41       & 45.99       & 49.33          & 57.62          & 59.63                   & 59.63                \\
    \textbf{Perceptron}      & 53.34       & 51.47       & 52.67          & 58.29          & 56.68                   & 57.09                \\ \hline
    \end{tabular}
    \caption{Accuracy of our sentiment analysis experiment results on scale of 0 to 100.}
    \label{tab:sentiment-experiment}
\end{table}

\section{Conclusion} 
\label{sec:conclusion}
In this paper, we described the creation of a fairly large narrow-domain corpus of Twitter posts related to the topic of eating. We gave some insights in overall observations gained from the corpus contents and various trends that we noticed from the data. We believe that the data would be useful in many linguistic, sociological, behavioural and other research areas.

We experimented with creating a food-related question answering system using one subset of our data and a sentiment analysis system using another subset to highlight potential use-cases of our corpus. While the results did not break new ground, we hope that they inspire related future research.


\section*{Acknowledgements}
\label{sec:acknowledgments}

We would like to thank Mārcis Pinnis for sharing his collected tweet dataset with us as well as running experiments with his model using our data.

\bibliographystyle{ios1}
\bibliography{Mendeley}

\end{document}